\documentclass[acmlarge, authorversion=true, nonacm=true]{acmart}
\usepackage{multirow}
\usepackage{float}
\usepackage{multicol}
\usepackage{array}
\usepackage{pifont}
\usepackage{soul}

\newcommand{\matr}[1]{\mathbf{#1}} 

\AtBeginDocument{%
  }
	
\begin{document}

\title[{Temporal Action Localization for Inertial-based Human Activity Recognition}]{Temporal Action Localization for Inertial-based Human Activity Recognition}

\author{Marius Bock}
\email{marius.bock@uni-siegen.de}
\orcid{1234-5678-9012}
\affiliation{%
  \institution{University of Siegen}
  \department{Ubiquitous Computing}
  \department{Computer Vision}
  \city{Siegen}
  \country{Germany}
}

\author{Michael Moeller}
\orcid{0000-0002-0492-6527}
\email{michael.moeller@uni-siegen.de}
\affiliation{%
  \institution{University of Siegen}
  \department{Computer Vision}
  \city{Siegen}
  \country{Germany}
}

\author{Kristof Van Laerhoven}
\orcid{0000-0001-5296-5347}
\email{kvl@eti.uni-siegen.de}
\affiliation{%
  \institution{University of Siegen}
  \department{Ubiquitous Computing}
  \city{Siegen}
  \country{Germany}
}

\begin{abstract}
As of today, state-of-the-art activity recognition from wearable sensors relies on algorithms being trained to classify fixed windows of data. In contrast, video-based Human Activity Recognition, known as Temporal Action Localization (TAL), has followed a segment-based prediction approach, localizing activity segments in a timeline of arbitrary length. This paper is the first to systematically demonstrate the applicability of state-of-the-art TAL models for both offline and near-online Human Activity Recognition (HAR) using raw inertial data as well as pre-extracted latent features as input.  Offline prediction results show that TAL models are able to outperform popular inertial models on a multitude of HAR benchmark datasets, with improvements reaching as much as 26\% in F1-score. We show that by analyzing timelines as a whole, TAL models can produce more coherent segments and achieve higher NULL-class accuracy across all datasets. We demonstrate that TAL is less suited for the immediate classification of small-sized windows of data, yet offers an interesting perspective on inertial-based HAR -- alleviating the need for fixed-size windows and enabling algorithms to recognize activities of arbitrary length. With design choices and training concepts yet to be explored, we argue that TAL architectures could be of significant value to the inertial-based HAR community. The code and data download to reproduce experiments is publicly available via \url{github.com/mariusbock/tal_for_har}.
\end{abstract}




\maketitle

\section{Introduction}
The recognition of performed activities through wearable sensors such as Inertial Measurement Units (IMUs) has shown to be of significant value in areas such as health care or the support of complex work processes \cite{bullingTutorialHumanActivity2014}. With the works of many researchers exploring lightweight architectures \cite{zhouTinyHARLightweightDeep2022}, much of the success of inertial-based models has stemmed from applying them in an online-fashion on wearable edge devices \cite{raviActivityRecognitionAccelerometer2005, wardEvaluatingPerformanceContinuous2006}. As shown to work on a multitude of HAR-related scenarios, it has established itself within the inertial-based community to employ a window-based prediction approach using a predefined window size and overlap. With classification algorithms being tasked to assign a label to each window individually, this approach has enabled inertial-based architectures to classify newly incoming data at will. As a too large window size may include multiple fast actions within a single window, and a too short window may not be enough to capture a complete lengthier action, the chosen size of the fixed window has become a crucial parameter within traditional recognition systems.

A persistent trend in Deep Learning has been the applicability of machine learning concepts such as self-attention \cite{vaswaniAttentionAllYou2017} to other areas and application scenarios than originally introduced for. With significant progress having been made since the introduction of deep neural architectures such as DeepConvLSTM \cite{ordonezDeepConvolutionalLSTM2016}, researchers have followed this trend and have continuously worked on improving the architectural design of networks by incorporating newly introduced techniques (see, e.g., \cite{zhouTinyHARLightweightDeep2022}). A promising recent approach in video-based Human Activity Recognition (HAR) is Temporal Action Localization (TAL), which aims to locate activity segments, defined by a class label, start, and end point, within an untrimmed video. Even though introduced architectures have almost doubled in performance over the last 5 years on existing datasets like THUMOS-14 \cite{jiangTHUMOSChallengeAction2014}, results on large corpora such as EPIC-KITCHENS-100 \cite{damenRescalingEgocentricVision2022} and Ego4D \cite{graumanEgo4DWorld0002022} show that the prediction problem is far from being saturated. 

Despite sharing a mutual goal, the TAL community as opposed to the inertial community does not follow an online-based prediction approach, but aims to classify recorded times sequences in an offline fashion, taking advantage of learnable temporal dependencies along the complete timeline. Furthermore, by following a segment-based prediction with activity boundaries being learned via regression-based loss, the "window size problem" is alleviated with methods being flexible enough to recognize both short and long lasting activities. TAL models have recently been shown to be capable of being trained using raw inertial data \cite{bockWEAROutdoorSports2023}, marking the first instance of such vision-based models being applied in the context of inertial-based HAR. When compared with popular inertial models, results on one particular dataset have already shown that TAL models can produce more coherent and less fragmented predictions while maintaining performance in terms of traditional classification metrics. In light of activity recognition systems having also been deployed to provide offline analysis of streams of prerecorded data \cite{inoueIntegratingActivityRecognition2019, vanlaerhovenUsingRhythmAwareness2008, debesMonitoringActivitiesDaily2016, blankeDailyRoutineRecognition2009, zhaiUbiSleepNetAdvancedMultimodal2021, berchtoldActiServActivityRecognition2010}, this work sets out to further examine the applicability of Temporal Action Localization for inertial-based HAR. Our contributions in this paper are three-fold:

\begin{enumerate}
    \item We demonstrate the capabilities of a novel approach to inertial-based HAR, being single-stage TAL, to outperform popular, window-based inertial models on a multitude of wearable activity recognition benchmark datasets in an offline-prediction scenario.
    \item Complementing traditional, window-wise classification metrics, we introduce a set of unexplored, segment-based evaluation metrics for inertial-based HAR, which are based around the scalar evaluation metric mean Average Precision \cite{alwasselDiagnosingErrorTemporal2018}.
    \item Though our results demonstrate the superiority of inertial-based models in an online prediction scenario, we show that TAL models can be applied in a near-online fashion, functioning e.g. a server-side prediction tool which provides prediction with a lag of 1 minute.
\end{enumerate}

\section{Related Work}
\paragraph{Inertial-based Human Activity Recognition}
With on-body sensors providing a robust and non-intrusive way to monitor participants along long stretches of time, research conducted in the area of inertial-based HAR has worked towards the automatic interpretation of one or multiple sensor streams to reliably detect activities e.g. in the context of providing medical support or providing guidance during complex work processes \cite{bullingTutorialHumanActivity2014}. With deep neural networks having established themselves as the de facto standard in inertial-based HAR, DeepConvLSTM \cite{ordonezDeepConvolutionalLSTM2016} as well as the models introduced at a later point follow a similar prediction scenario design applying a sliding window approach which groups concurrent data points for classification. In their original publication \citet{ordonezDeepConvolutionalLSTM2016} have found a combination of convolutional and recurrent layers to produce competitive results, with the idea being to model temporal dependencies amongst automatically extracted discriminative features within a sliding window in order to classify it correctly as one of the $N$ activity classes or, if applicable, NULL-class. Building on the idea of combining these two types of layers, researchers have worked on extending the original DeepConvLSTM or have introduced their own architectural designs \cite{murahariAttentionModelsHuman2018, vaizmanContextRecognitionInthewild2018, pengAROMADeepMultitask2018, xiDeepDilatedConvolution2018, xuInnoHARDeepNeural2019, xiaLSTMCNNArchitectureHuman2020, liuGIobalFusionGlobalAttentional2020, chenMETIERDeepMultitask2020, bockImprovingDeepLearning2021, abedinAttendDiscriminateStateoftheart2021, zhouTinyHARLightweightDeep2022, zhangIFConvTransformerFrameworkHuman2022, miaoDynamicIntersensorCorrelations2022, chenSALIENCEUnsupervisedUser2023, chenHMGANHierarchicalMultimodal2023, wielandTinyGraphHAREnhancingHuman2023}. A simple modification of the DeepConvLSTM is the shallow DeepConvLSTM \cite{bockImprovingDeepLearning2021}. Contradicting the popular belief that one needs at least two recurrent layers when dealing with time series data \cite{karpathyVisualizingUnderstandingRecurrent2015}, \citet{bockImprovingDeepLearning2021} demonstrated that removing the second LSTM layer within the original DeepConvLSTM architecture results in significant increases in predictive performance on a multitude of HAR benchmark datasets while also decreasing the number of learnable parameters. Furthermore, with the original DeepConvLSTM only being able to learn temporal dependencies within a sliding window, the shallow DeepConvLSTM applies the LSTM across batches, effectively making the batches the sequence which is to be learned by the LSTM. This dimension flip, along with a non-shuffled training dataset, enables the architecture to learn temporal dependencies amongst a batched input. The same year as the shallow DeepConvLSTM, \citet{abedinAttendDiscriminateStateoftheart2021} introduced Attend-and-Discriminate, a deep neural network architecture following the idea of the original DeepConvLSTM by combining both convolutional and recurrent layers, yet further adding a cross-channel interaction encoder using self-attention as well as attention mechanism to the recurrent parts of the network. In 2022 \citet{zhouTinyHARLightweightDeep2022} proposed TinyHAR, a lightweight HAR model that uses a transformer encoder block along with means of cross-channel fusion via a fully connected layer and a final self-attention layer which aims to learn the contribution of each outputted time step produced by the recurrent parts individually.

\paragraph{Video-based Human Activity Recognition}
Classifying videos in the context of Human Activity Recognition can be broadly categorized into three main application scenarios: Action Recognition, which aims to classify trimmed videos into one of $C$ activity classes; Action Anticipation, which aims to predict the next likely activities after observing a preceding video sequence; and Temporal Action Localization (TAL), which seeks to identify and locate all activity segments within an untrimmed video. With the inertial-based benchmark datasets consisting of a multitude of activities, TAL is to be considered most comparable to inertial-based HAR. Unlike popular intertial architectures though, TAL models aim to predict all segments within an untrimmed video at once. Existing TAL methods can broadly be categorized into two categories: two-stage and one-stage recognition systems. Two-stage recognition system \cite{linBMNBoundarymatchingNetwork2019, linFastLearningTemporal2020, xuGTADSubgraphLocalization2020, baiBoundaryContentGraph2020, zhaoBottomupTemporalAction2020, zengGraphConvolutionalNetworks2019, gongScaleMattersTemporal2020, liuMultishotTemporalEvent2021, qingTemporalContextAggregation2021, sridharClassSemanticsbasedAttention2021, zhuEnrichingLocalGlobal2021, zhaoVideoSelfstitchingGraph2021, tanRelaxedTransformerDecoders2021} divide the process of identifying actions within an untrimmed video into two subtasks. First, during proposal generation, candidate segments are generated, which are then, during the second step, classified into one of $C$ activity classes and iteratively refined regarding their start and end times. Contrarily, single-stage models \cite{yangBasicTADAstoundingRgbonly2023, shiReActTemporalAction2022, nagProposalfreeTemporalAction2022, liuEndtoendTemporalAction2022, liuProgressiveBoundaryRefinement2020, longGaussianTemporalAwareness2019, linLearningSalientBoundary2021, chenDCANImprovingTemporal2022, zhangActionFormerLocalizingMoments2022, tangTemporalMaxerMaximizeTemporal2023, shiTriDetTemporalAction2023} do not rely on activity proposals, directly aiming to localize segments along their class label and refined boundaries. In 2022 \citet{zhangActionFormerLocalizingMoments2022} introduced the ActionFormer, a lightweight, single-stage TAL model which unlike previously introduced single-stage architectures does not rely on pre-defined anchor windows. In line with the success of transformers in other research fields, \citet{zhangActionFormerLocalizingMoments2022} demonstrated their applicability for TAL, outperforming previously held benchmarks on several TAL datasets \cite{heilbronActivityNetLargescaleVideo2015, damenRescalingEgocentricVision2022, jiangTHUMOSChallengeAction2014} by a significant margin. Surprisingly, a year later, TemporalMaxer suggested removing transformer layers within the ActionFormer, arguing that feature embeddings are already discriminative enough \cite{tangTemporalMaxerMaximizeTemporal2023}. Though being more lightweight than the ActionFormer, the TemporalMaxer showed to outperform its precedent during benchmark analysis. Similarly to the TemporalMaxer, \citet{shiTriDetTemporalAction2023} introduced TriDet, which suggested altering ActionFormer in two ways. First, to mitigate the risk of rank-loss, self-attention layers are replaced with SGP layers. Second, the regression head in the decoder is replaced with a Trident head, which improves imprecise boundary predictions via an estimated relative probability distribution around each timestamp's activity boundaries.

\section{Temporal Action Localization for Inertial-based HAR}
\label{sec:talforhar}
\begin{figure*}
\centering
\includegraphics[width=1.0\textwidth]{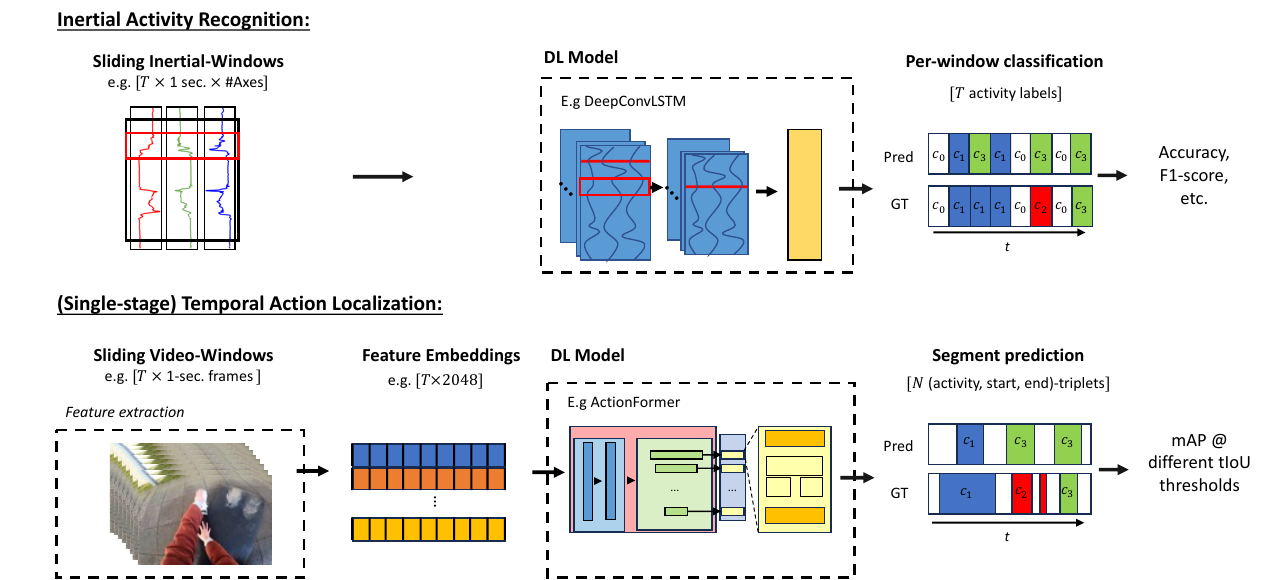}
    \caption{Overview of the prediction pipelines applied in inertial-based activity recognition and single-stage Temporal Action Localization (TAL). Both apply a sliding window to divide input data into windows of a certain duration (e.g. one second). TAL models do not use raw data as input but are applied on per-clip, pre-extracted feature embeddings. Inertial activity recognition models predict activity labels for each sliding window, which are used for calculating classification metrics such as accuracy and F1. TAL models predict activity segments, defined by a label, start and end points, and are evaluated with mean Average Precision (mAP) applied at different temporal Intersection over Union (tIoU) thresholds.}
    \label{fig:inertialvsvideo}
    \Description{A visualization of the prediction pipeline of inertial-based activity recognition and single-stage Temporal Action Localization. Within inertial-based activity recognition inertial data, illustrated as three sensor streams, is chunked into sliding windows which are then predicted as one of the activity classes by a Deep Learning model. The predicted labels are then compared to that of the ground truth and classification metrics are calculated. Temporal Action Localization first extracts video embeddings from sliding video clips. The embeddings are then fed through a Deep Learning model which predicts activity segments, visualized as colored boxes on a timeline, that are in a final step compared for overlap and label correctness with the ground truth.}
\end{figure*}

As the inertial-based HAR and TAL communities deal with inherently different modalities, both communities have developed distinct predictive pipelines and algorithms tailored to the challenges and characteristics of their respective modalities (see Figure \ref{fig:inertialvsvideo}). The objective of both inertial activity recognition and TAL is to predict all activities present in an untrimmed timeline. Given an input data stream $X$ of a sample participant, both the inertial and TAL communities start by applying a sliding window approach which shifts over $X$, dividing the input data into windows, e.g. of one second duration with a 50\% overlap between consecutive windows. This process results in $X = \{\matr{x}_1, \matr{x}_2, ..., \matr{x}_T\}$ being discretized into $t = \{0, 1, ..., T\}$ time steps, where $T$ is the number of windows for each participant. It is important to note that the TAL models do not use raw data as input but are instead trained using feature embeddings extracted from each individual sliding video clip, which are extracted using pre-trained methods such as \cite{carreiraQuoVadisAction2017}.

Given all sliding windows associated with an untrimmed sequence, inertial activity recognition models aim to predict an activity label $a_t$ for each sliding window $\matr{x}_t$, where $a$ belongs to a predefined set of activity labels, $a_t \in {1, ..., C}$. To do so, the sliding windows are batched and fed through, e.g., a deep neural network, such as the DeepConvLSTM \cite{ordonezDeepConvolutionalLSTM2016}. The resulting activity labels for each window are then compared to the true activity labels from the ground truth data, and classification metrics like accuracy or F1-score are calculated. Contrarily, TAL models aim to identify and localize segments of actions within the untrimmed data stream, which can span across multiple windows. To achieve this algorithms are trained to predict activity segments $Y = \{\matr{y}_1, \matr{y}_2, ..., \matr{y}_N\}$, where $N$ varies across participants. Each activity instance $\matr{y_i} = (s_i, e_i, a_i)$ is defined by its starting time $s_i$ (onset), end time $e_i$ (offset) and its associated activity label $a_i$, where $s_i \in [0,T]$, $e_i \in [0,T]$, and $a_i \in \{1, ..., C\}$. 

As the input data used to train TAL models is a collection of 1-dimensional feature embeddings, the 2 d-dimensional, windowed inertial data commonly found in the inertial-based HAR community, cannot directly be used to train TAL models. The following will thus describe two preprocessing methods which can be used to convert the inertial data into a format such that it can be used as input for TAL models.

\begin{figure*}
\centering
\includegraphics[width=0.8\textwidth]{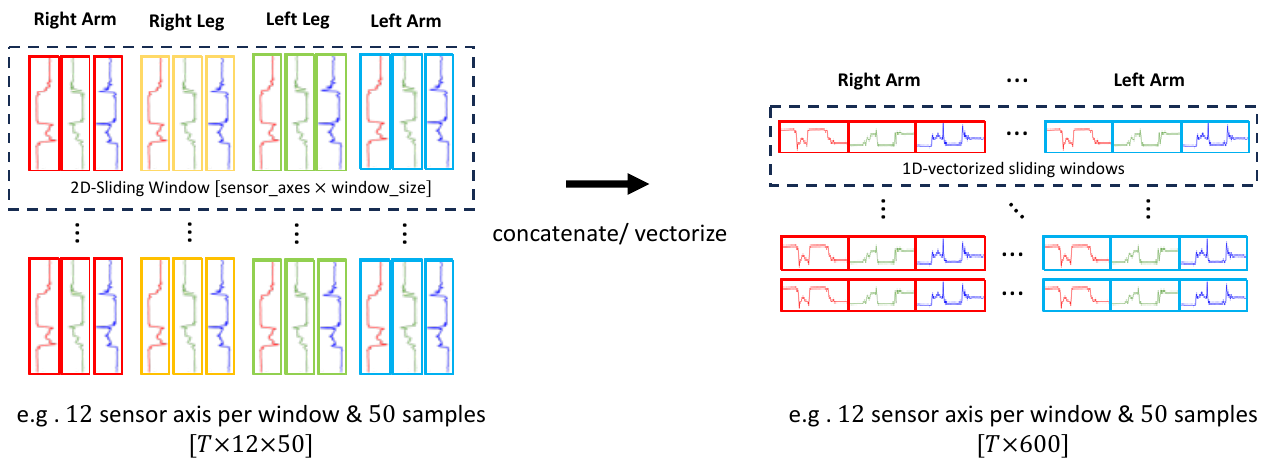}
    \caption{Visualization of the applied vectorization on top of windowed inertial data assuming four 3D-inertial sensors and a sliding window size of 50 samples. Each 2D-sliding-window of size $[50 \times 12]$ is vectorized by concatenating each of the axes one after another. Resulting 1D-embedding vectors, being of size $[1 \times 600]$, can be used to train TAL models.}
    \label{fig:vectorization}
    \Description{Visualization of the vecotrization of a sliding window of the inertial data such that it can be used as input for TAL models. For each window each of the three axis of a sensors are grouped together (i.e., a 2-dimensional matrix). The vectorization applied in this paper concatenates each axis of each sensors one after another such that the each two-dimensional sliding window is converted to a 1-dimensional feature embedding. Within the example 12 axis with 50 samples each are thus converted to a feature vector of size 600.}
\end{figure*}

\paragraph{Vectorization of raw inertial data}
Since both communities employ a sliding window approach but feed data to their models using different dimensions, \citet{bockWEAROutdoorSports2023} proposed a simple, yet effective preprocessing step. This step converts the 2-dimensional, windowed inertial data, as used by inertial architectures, into a 1-dimensional feature embedding suitable for training TAL models. The preprocessing method as detailed in Figure \ref{fig:vectorization} involves concatenating the different sensor axes of each window, converting the input data to be a collection of 1-dimensional feature embedding vectors $\matr{x}_t\in \mathbb{R}^{1 \times W S}$ where $W$ is the number of samples within a window, and $S$ is the number of sensor axes. More specifically, given a sliding window $\matr{x}_{SW} \in \mathbb{R}^{W \times S}$, we vectorize the two-dimensional matrix as follows:
\begin{equation}
\label{eq:vectorization}
\matr{x}_{SW} = \begin{bmatrix} 
    x_{11} & \dots & x_{1S} \\
    \vdots & \ddots & \\
    x_{W1} &        & x_{WS}
    \end{bmatrix}
\to
\text{vec}(\matr{x}_{SW}) = \begin{bmatrix} 
    x_{11} \\
    \vdots \\ 
    x_{1S} \\ 
    x_{2S} \\ 
    \vdots \\
    x_{WS}
    \end{bmatrix}
\end{equation}

\begin{figure*}
\centering
\includegraphics[width=0.9\textwidth]{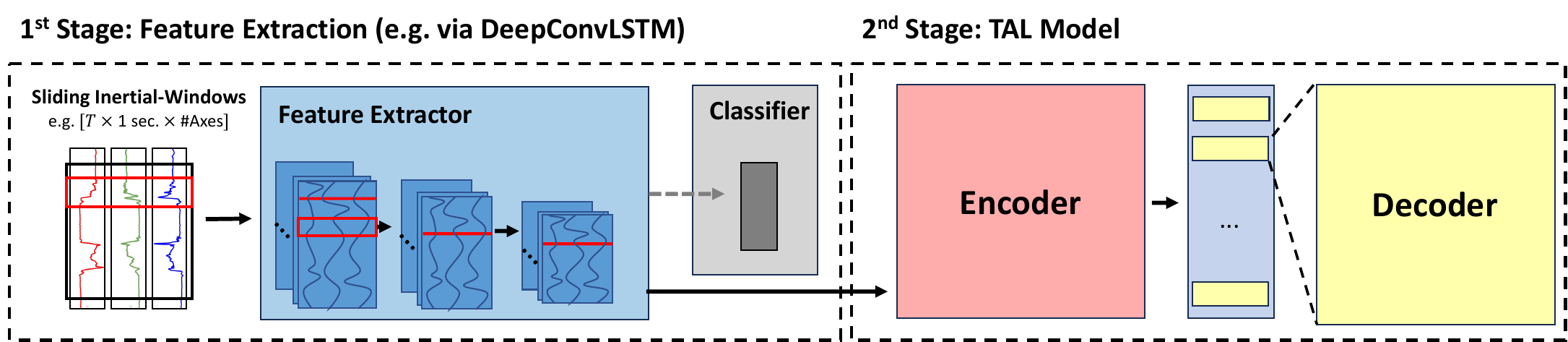}
    \caption{Visualization of the applied two-stage training process. The first stage involves training e.g. a classic DeepConvLSTM as introduced by \citet{ordonezDeepConvolutionalLSTM2016}. Once the first-stage training has finished, the classifier is omitted from the model such that latent features can be extracted. The 1-dimensional, window-wise features are then used as input embeddings for the second stage, i.e. training a TAL model.}
    \label{fig:twostage}
    \Description{Visualization of the applied two-stage training process. The first stage involves training e.g. a classic DeepConvLSTM as introduced by \citet{ordonezDeepConvolutionalLSTM2016}. Once the first-stage training has finished, the classifier is omitted from the model such that latent features can be extracted. The 1-dimensional, window-wise features are then used as input embeddings for the second stage, i.e. training a TAL model.}
\end{figure*}

\paragraph{Two-stage training via prepended inertial models} In order to encode videos and come up with a discriminative, latent feature representation, TAL models usually resort to using extracted feature embeddings from models pretrained on large vision corpuses like Kinetics-400 \cite{kayKineticsHumanAction2017}. Inspired by this, we propose a second variant on how to use inertial data as input to TAL models (see Figure \ref{fig:twostage}) which involves using features extracted from a separately trained inertial model as latent representation of each sliding window. Specifically, a LOSO cross-validation step within the two stage training consists of 1. training the inertial model on all but the validation data, 2. use the trained inertial network to extract latent feature representations of each window within the training and validation data, and 3. using the extracted features as clip-wise feature embeddings to train a TAL model.

\subsection{TAL Architectures Overview} 

\begin{figure*}
\centering
\includegraphics[width=0.9\textwidth]{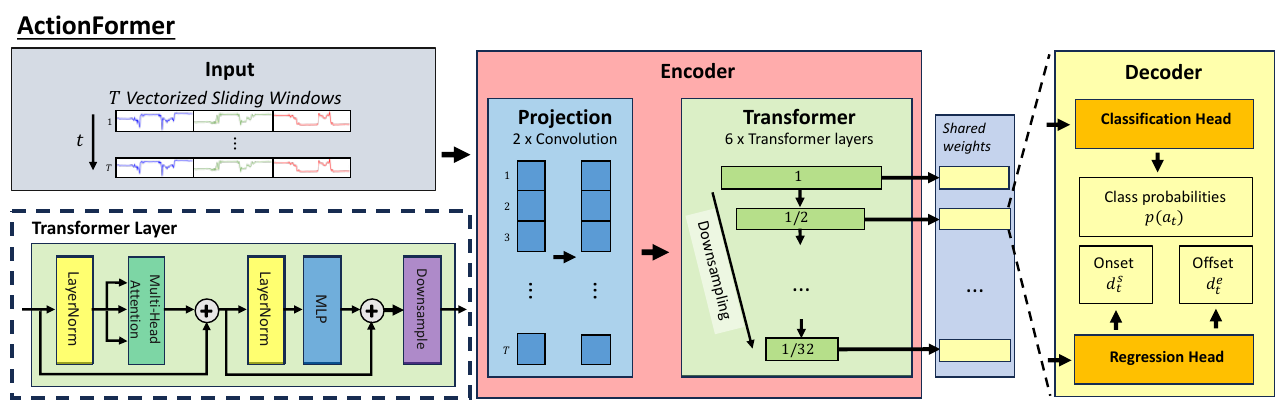}
    \caption{Architecture overview of the ActionFormer proposed by \citet{zhangActionFormerLocalizingMoments2022}. The architecture follows a encoder-decoder structure. The encoder encodes input sequences into a feature pyramid, which captures information at various temporal scales. The decoder, consisting of a classification and regression head, then decodes each timestamp within the feature pyramid to sequence labels, i.e. a class probability vector and the timestamp's activity onset and offset distance. The TriDet\cite{shiTriDetTemporalAction2023} and TemporalMaxer\cite{tangTemporalMaxerMaximizeTemporal2023} both follow the same encoder-decoder structure, yet suggest architectural changes.}
    \label{fig:actionformer}
    \Description{Visualization of the ActionFormer architecture. Inputs, i.e. 1-dimensional feature embeddings each representing a sliding window of a subject's data stream, are passed through an encoder which consists of projection and transformer layers. Each layer of the feature pyramid of the encoder is passed to a decoder which predicts class probabilities, onset and offset of each data point within the feature pyramid. Weights of the decoder are shared across the feature pyramid layers.}
\end{figure*}

In light of the recent success of transformer-based models in e.g. Natural Language Processing and Computer Vision, \citet{zhangActionFormerLocalizingMoments2022} proposed the \emph{ActionFormer}, is an end-to-end trainable transformer-based TAL architecture. Unlike other single-stage TAL approaches, it does not rely on pre-defined anchor windows. The architecture, as illustrated in Figure~\ref{fig:actionformer}, combines multiscale feature representations with local self-attention and is trained through a classification and regression loss calculated by a light-weighted decoder. Building up on the ActionFormer architecture, \citet{tangTemporalMaxerMaximizeTemporal2023} and \citet{shiTriDetTemporalAction2023} proposed the \emph{TemporalMaxer} and \emph{TriDet} model respectively. Within the \emph{TriDet} model projection and transformer layers of the ActionFormer are replaced with fully-convolutional SGP layers and the regression head is replaced by a trident regression head which claims to improve imprecise boundary predictions. Contrarily, the TemporalMaxer suggests modifying the encoder of the ActionFormer to employ solely max pooling and remove all transformer-based layers, as, according to the authors, this does not come at the cost of a lost in information and predictive performance.

In order to predict activity segments $Y = \{\matr{y}_1, \matr{y}_2, ..., \matr{y}_N\}$ within an input video, the ActionFormer, TemporalMaxer and TriDet model all follow the same sequence labeling problem formulation for action localization. That is, given a set of feature input vectors $X = \{\matr{x}_1, \matr{x}_2, ..., \matr{x}_T\}$, a model aims to classify each timestamp as either one of the activity categories $C$ or as background (or null) class and regress the distance towards the timestamp's corresponding segment's start and end point. More specifically, given the input vectors $X$ a model aims to learn to label the sequence as

\begin{equation}
\label{eq:seqlabel}
X = \{\matr{x}_1, \matr{x}_2, ..., \matr{x}_T\} \to \matr{\hat{Y}} = \{\matr{\hat{y}_1}, \matr{\hat{y}_2}, ... ,\matr{\hat{y}_T} \} ,
\end{equation}
where $\matr{\hat{y}_t} = (p(a_t), d_t^s, d_t^e)$ at timestamp $t$ is defined by a probability vector $p(a_t)$ indicating the class-wise probability of the timestamp being classified as one of the activity categories $C$ and $d_t^s > 0$ and $d_t^e > 0$ being the distance between the current timestamp $t$ and the current segment's onset and offset. Note that $d_t^s$ and $d_t^e$ are not defined if the timestamp is to be classified as background. 
The sequence labeling formulation can then be easily decoded to activity segments with: 
\begin{equation}
  a_t = \arg \max p(a_t), \quad s_t = t - d_t^s, \quad \textrm{and} \quad e_t = t + d_t^e
\end{equation}

The authors of the ActionFormer, TemporalMaxer, and TriDet models attribute much of their models' performance to the constructed multi-layer feature pyramid. The feature pyramid within each of the three models downsamples the input sequence of sliding windows multiple times to create representations of a participant's data stream at different temporal granularities \cite{zhangActionFormerLocalizingMoments2022, tangTemporalMaxerMaximizeTemporal2023, shiTriDetTemporalAction2023}. Using this method, the TAL models can learn both short and long temporal dependencies across a participants's timeline, as the temporal distance between two embedding vectors within the feature pyramid increases the further we move down the pyramid. This use of downsampling to capture different length temporal patterns is comparable to how inertial models concatenate convolutional layers without padding to increase the receptive field of each convolutional kernel. However, both types of models differ in the types of patterns learned. The feature pyramid of the TAL models enables them to learn cross-window, temporal patterns of arbitrary length, while the convolutional parts of the inertial models enable them to learn in-window temporal patterns of arbitrary length.

Due to the TAL community's training objective differing from that of the inertial community, TAL models are trained not only on a classification loss, but also a regression loss, which optimizes each timestamp's corresponding activity onset and offset. The use of self-attention based layers has shown to improve results in both communities, yet variations of the ActionFormer show that these layers are not necessarily needed. More specifically, the use of fully-convolutional SGP layers in the TriDet model show that, rather than the type of technique being important, focus needs to be put specifically on learning both local and global temporal information for each timestamp.

\section{Methodology}
\subsection{Datasets}
We evaluate each algorithm featured in this benchmark analysis using 6 popular HAR datasets, namely the Opportunity \cite{roggenCollectingComplexActivity2010}, SBHAR \cite{reyes-ortizTransitionAwareHumanActivity2016}, Wetlab \cite{schollWearablesWetLab2015}, WEAR \cite{bockWEAROutdoorSports2023}, Hang-Time \cite{hoelzemannHangtimeHARBenchmark2023} and RWHAR dataset \cite{sztylerOnbodyLocalizationWearable2016}. The datasets, all covering different application and recording scenarios, provide us with a challenging set of prediction problems to properly assess the strengths and weaknesses of each model. Table \ref{tab:datasets} summarizes the key characteristics of each dataset. In addition to vital information such as participant count, activity count, sampling rate and sensor axes count, the table also includes details on the scenario and type of activities found in each dataset. 
We classify activities into four types: (1) periodic activities, characterized by recurring periodic patterns; (2) non-periodical activities consisting of non-occurring, non-periodical patterns; and (3) complex activities, defined by an arbitrary sequence of non-periodic and periodic activities. Lastly, following the works of \citet{alwasselDiagnosingErrorTemporal2018}, we provide average count of segments across all subjects, categorizing each segment within each dataset into five bins: XS: (0 seconds, 3 seconds], S: (3 seconds, 6 seconds], M: (6 seconds, 12 seconds], L: (12 seconds, 18 seconds], and XL: more than 18 seconds.

\begin{table}
\scriptsize
\caption{Key characteristics of the datasets used in this benchmark analysis. The table provides: participant count (\#Sbjs), activity count (\#Cls), sampling rate (SR), sensor axes count (\#Axes), average number of segments per participant based on absolute length of the segment (\#Segments), overall scenario (e.g. activities of daily living (ADL)) and type of activities found in the dataset.}
\label{tab:datasets}
{
\begin{tabular}{lccccccccccl}
 Dataset        & \#Sbjs & \#Cls & SR & \#Axes  & \multicolumn{5}{c}{\#Segments \cite{alwasselDiagnosingErrorTemporal2018}}  & Scenario & Type of Activities \\
                &        &       & &              & XS & S & M & L & XL        &                    \\
 \toprule
 Opportunity \cite{roggenCollectingComplexActivity2010}     & 4  & 17(+1) & 30 & 113 & $403\pm63$ & $185\pm48$ & $48\pm10$ & $3\pm2$ & $0\pm0$ & ADL & non-periodic, complex \\
 SBHAR \cite{reyes-ortizTransitionAwareHumanActivity2016}   & 30 & 12(+1) & 50 & 3 & $3\pm2$ & $9\pm2$ & $6\pm4$ & $10\pm4$ & $13\pm3$ & Locomotion & (non-)periodic\\
 Wetlab \cite{schollWearablesWetLab2015}                    & 22 & 8(+1) & 50 & 3 & $7\pm5$ & $7\pm3$ & $7\pm2$ & $5\pm2$ & $13\pm2$ & Laboratory & complex\\
 WEAR \cite{bockWEAROutdoorSports2023}                      & 18 & 18(+1) & 50 & 12 & $1\pm1$ & $1\pm1$ & $1\pm1$ & $1\pm1$ & $30\pm9$ & Sports & periodic\\
 Hang-Time \cite{hoelzemannHangtimeHARBenchmark2023}         & 24 & 5(+1) & 50 & 3 & $154\pm53$ & $22\pm7$ & $10\pm7$ & $3\pm3$ & $3\pm2$ & Sports & (non-)periodic, complex\\
 RWHAR \cite{sztylerOnbodyLocalizationWearable2016}         & 15 & 8 & 21 & 50 & $0\pm0$ & $0\pm0$ & $0\pm0$ & $0\pm0$ & $9\pm1$ & Locomotion & periodic \\
\end{tabular}
}
\end{table}

\subsection{Training Pipeline}
\paragraph{Prediction Scenarios}
All experiments were initiated employing a Leave-One-Subject-Out cross-validation. This type of validation involves iteratively training on all but one participant's data and using the hold-out participant's data during validation until all participants have been evaluated, ensuring that each network architecture is assessed based on its capabilities to generalize across unseen participants. All datasets were windowed using a sliding window of one second with a 50\% overlap across windows. In order to allow for fair comparison, our benchmark analysis assesses the TAL models employing two prediction scenarios, with the differences in their predicted output between inertial-based and TAL models. Being the scenario TAL models were intended to be used for, within the first prediction scenario (\textit{Offline Activity Recognition}) the TAL models are tasked to predict each participant's data as a whole, i.e. one batch representing the full data stream available of one participant. Nevertheless, to also assess the models' capabilities to provide window-level predictions, within the second prediction scenario (\textit{Online Activity Recognition}) we artificially chunk participants' timelines making the models predict the timelines in chunks (e.g., 30 seconds). As inertial-based models unlike TAL models are designed to make window-based predictions, the second prediction scenario will assess the TAL models online capabilities while also benchmarking them with removed temporal context. Note that in case of the \textit{Offline Activity Recognition} we test the TAL models using both a single-stage as well as two-stage training (as described in Chapter~\ref{sec:talforhar}).
 
\paragraph{Hyperparameters} 
For all inertial architectures \cite{ordonezDeepConvolutionalLSTM2016, bockImprovingDeepLearning2021, abedinAttendDiscriminateStateoftheart2021, zhouTinyHARLightweightDeep2022} we employed a similar optimization as proposed with the release of the shallow DeepConvLSTM, namely an Adam optimizer paired with a learning rate of $1e^{-4}$, a weight decay of $1e^{-6}$, and Glorot weight initialization \cite{glorotUnderstandingDifficultyTraining2010}. To allow each model to converge more properly, we increase the number of epochs to 100 and employ a step-wise learning-rate schedule that multiplies the learning rate by a factor of 0.9 after every 10 epochs. For all architectures, we fixed the hidden dimension of the recurrent layers to employ 128 units and scaled kernel sizes of the convolutional filters according to the relative difference in sampling rate among the different input datasets. In line with how the Attend-and-Discriminate architecture was first introduced, we optimized said architecture using center-loss \cite{wenDiscriminativeFeatureLearning2016} as opposed to a weighted cross-entropy loss, which was used during training of all other inertial architectures. Lastly, as proposed by the authors, we do not shuffle batches during the training of the shallow DeepConvLSTM.

The three TAL architectures \cite{zhangActionFormerLocalizingMoments2022, tangTemporalMaxerMaximizeTemporal2023, shiTriDetTemporalAction2023}, we chose to employ hyperparameters reported by the authors that produced best results on the EPIC-Kitchens dataset \cite{damenRescalingEgocentricVision2022}. The hyperparamenters, though optimized for a different modality than inertial data (egocentric videos), have shown to produce competitive results on the WEAR dataset \cite{bockWEAROutdoorSports2023} and are thus considered a good starting point for evaluating the applicability of the three architectures on other HAR datasets. Nevertheless, given the small size of the tested inertial datasets compared to datasets used by the TAL community, we chose to increase the amount of epochs to 100 throughout all TAL-based experiments. As TAL models are designed to predict both a regression (boundary prediction) and classification target (segment label), loss calculation of the ActionFormer, TemporalMaxer and TriDet models are performed as a weighted combination of a regression loss (IoU loss \cite{rezatofighiGeneralizedIntersectionUnion2019}) and classification loss (focal loss \cite{liGeneralizedFocalLoss2020}).

\paragraph{Postprocessing}
To compare predictions of both TAL and inertial-based models, segments (TAL) and windowed-predictions (inertial-based) first needed to be translated back to sample-wise predictions. In case of the segments we do so by translating each segment "overwrite" the prediction timeline with its associated label. By starting with the segment associated with the lowest prediction confidence, segments with a high prediction confidence are preferred in case of overlapping segments. In case of the inertial-based architectures, we start by iterating over the windowed predictions in order of occurrence in the timeline having them determine the samples they are associated with. To deal with the overlap amongst windows, preceding windows are allowed to "overwrite" the label of samples of previous windows which they are overlapping with.

As inertial models are tasked to predict on a per-window basis and, the architectures suffer from frequently occurring activity switches ultimately leading to fragmented segments which significantly lower mAP scores being produced as opposed to the TAL models. Therefore, to remove only short lasting switches, predictions of inertial-based architectures mentioned in this paper were smoothed using majority-vote filters. The exact size of the majority filter was chosen dataset-specific, determined via trying out a selection of filters between 2.5 and 40 seconds. Specifically, the filters were chosen as: 2.5 seconds (Opportunity), 20 seconds (Wetlab), 5 seconds (SBHAR), 15 seconds (WEAR), 5 seconds (Hang-Time) and 40 seconds (RWHAR). Similar to the inertial models, the TAL architectures are tasked to predict class probabilities and segment boundaries of each windowed timestamp. Consequently, without applying any confidence threshold, all predicted activity segments are considered during creation of the prediction timeline causing accuracy of the NULL-class to be significantly low. Therefore, to alleviate this, we apply an optimized confidence threshold on predicted segments of all TAL models. Similarly to the majority filter, the score threshold for each architecture was chosen dataset-specific, determined via trying out a selection of thresholds between 0.05 and 0.5 seconds. This eliminates low scoring segments and substantially lowers the confusion of the NULL-class with the other activities. More details on the effect of the majority filter as well as score thresholding can be found in the supplementary material.

\section{Results}
As part of our experimental evaluation, we provide traditional classification metrics (precision, recall and F1-score), misalignment measures as defined by \citet{wardEvaluatingPerformanceContinuous2006} and mAP averaged across tIoU thresholds 0.3, 0.4, 0.5, 0.6 and 0.7. All results are the unweighted average across all subjects along the LOSO validation. Experiments were repeated three times employing three different random seeds (1, 2 and 3). Classification metrics are calculated on a per-sample basis as the segmented output of the TAL models and windowed output of the inertial-models need to be translated to a common time granularity. To ensure readability of this work, visualization of the per-class analysis will only include confusion matrices of the SBHAR and RWHAR datasets, as we deemed these two datasets to be the most representative in illustrating the strengths and weaknesses of the TAL architectures when applied to inertial data. Please note that the confusion matrices of the other datasets can be found in the supplementary material. Furthermore, all created plots part of a performed DETAD analysis \cite{alwasselDiagnosingErrorTemporal2018} on each dataset can be found in our repository.

\subsection{Offline Activity Recognition}

Table \ref{tab:perdataset} provides average results of the seven tested architectures across each dataset in an offline prediction setting. One can see that the TAL architectures outperform the inertial architectures across all datasets regarding average mAP. This shows that by being trained to specifically optimize activity boundaries, the different prediction target has resulted in overall more coherent segments which overlap to a larger degree with the ground truth segments. Even though prediction results of the inertial architectures were further smoothed by a majority filter, average mAP is, except for the WEAR dataset, more than halved when compared to that of the TAL architectures. Regarding traditional classification metrics the TAL architectures are able to outperform inertial architectures for four out of six datasets with only the RWHAR \cite{sztylerOnbodyLocalizationWearable2016} and WEAR dataset \cite{bockWEAROutdoorSports2023} being the exception. These improvements range between $5\%$ for the Opportunity and Wetlab dataset, $10\%$ for the Hang-Time and even as much as 25\% in F1-score for the SBHAR dataset. Calculated misalignment ratios show that both inertial and TAL architectures have a similar distribution of errors, with architectures producing better overall classification results also showing overall lower misalignment measures. Only for the RWHAR dataset one can see that TAL architectures show a significantly higher Overfill-Ratio. This might be due to the RWHAR dataset not featuring a NULL-class, which introduces an uncommon prediction scenario for the TAL architectures. Nevertheless, the performed DETAD analysis \cite{alwasselDiagnosingErrorTemporal2018}, which further differentiates amongst the segment-based errors, reveals that inertial architectures suffer more severely from background errors, i.e. confusing activities with the NULL-class. While this effect is decreased for the shallow DeepConvLSTM, TAL architectures show to be able to more reliably differentiate between activities and the NULL-class, and thus more reliably localize activities within the untrimmed sequences. 

\begin{table*}
  \centering
  \scriptsize
  \caption{\textit{Offline Activity Recognition:} Average LOSO cross-validation results obtained on six inertial HAR benchmark datasets \cite{roggenCollectingComplexActivity2010, reyes-ortizTransitionAwareHumanActivity2016, schollWearablesWetLab2015, bockWEAROutdoorSports2023, hoelzemannHangtimeHARBenchmark2023, sztylerOnbodyLocalizationWearable2016} for four inertial \cite{ordonezDeepConvolutionalLSTM2016, bockImprovingDeepLearning2021, abedinAttendDiscriminateStateoftheart2021, zhouTinyHARLightweightDeep2022} and three TAL architectures \cite{zhangActionFormerLocalizingMoments2022, tangTemporalMaxerMaximizeTemporal2023, shiTriDetTemporalAction2023}. The table provides per-sample classification metrics, i.e. Precision (P), Recall (R), F1-Score (F1), misalignment ratios \cite{wardEvaluatingPerformanceContinuous2006} and average mAP applied at different tIoU thresholds (0.3:0.1:07). All experiments employed a sliding window of one second with a $50\%$ overlap. Results are averaged across three runs employing different random seeds. The TAL architectures are able to outperform the inertial architectures regarding average mAP on all HAR datasets and result in the highest classification metrics on four out of the six datasets. Best results per dataset are in \textbf{bold}.}
  \label{tab:perdataset}
  \begin{tabular}{@{}llccc|cccccc|c@{}}
    & Model & P ($\uparrow$) & R ($\uparrow$) & F1 ($\uparrow$) & UR ($\downarrow$) & OR ($\downarrow$) & DR ($\downarrow$) & IR ($\downarrow$) & FR ($\downarrow$) & MR ($\downarrow$) & mAP ($\uparrow$) \\ 
    \hline
    \multirow{7}{*}{\rotatebox[origin=c]{90}{Opportunity}}
    & DeepConvLSTM              & 50.22 & 33.88 & 34.41 & 0.30 & 17.29 & 0.78 & \textbf{31.26} & \textbf{0.01} & 0.23 & 13.97 \\
    & Shallow D.                & 42.08 & 27.18 & 26.46 & 0.24 & 14.28 & 0.97 & 35.06 & \textbf{0.01} & \textbf{0.15} & 10.61 \\
    & A-and-D                   & 35.25 & 48.55 & 36.35 & 0.32 & 15.28 & 0.45 & 52.55 & 0.02 & 0.35 & 13.75 \\
    & TinyHAR                   & 48.09 & 54.27 & 47.09 & 0.34 & 19.99 & \textbf{0.39} & 34.01 & 0.02 & 0.46 & 19.78 \\ \cline{2-12}
    & ActionFormer              & \textbf{54.63} & \textbf{58.78} & \textbf{51.93} & \textbf{0.19} & 13.34 & 0.41 & 33.82 & \textbf{0.01} & 0.42 & \textbf{51.24} \\
    & TemporalMaxer             & 44.44 & 55.63 & 44.74 & 0.21 & 14.13 & 0.40 & 43.78 & \textbf{0.01} & 0.55 & 46.31 \\
    & TriDet                    & 48.72 & 57.69 & 48.79 & 0.23 & \textbf{13.20} & \textbf{0.39} & 40.07 & \textbf{0.01} & 0.58 & 49.70 \\
    \hline
    \multirow{7}{*}{\rotatebox[origin=c]{90}{SBHAR}}
    & DeepConvLSTM              & 67.54 & 63.72 & 62.31 & 0.41 & 7.19 & 0.59 & 21.44 & 0.07 & 0.10 & 49.60 \\
    & Shallow D.                & 72.98 & 75.41 & 71.13 & 0.60 & 10.19 & 0.46 & 14.23 & 0.02 & \textbf{0.09} & 65.15 \\
    & A-and-D                   & 68.64 & 71.07 & 66.49 & \textbf{0.31} & 9.47 & 0.45 & 21.71 & 0.05 & 0.11 & 55.79 \\
    & TinyHAR                   & 58.91 & 63.70 & 56.29 & \textbf{0.31} & 8.16 & 0.54 & 31.67 & 0.05 & 0.13 & 45.38 \\ \cline{2-12}
    & ActionFormer              & 87.02 & 84.37 & 84.43 & 0.36 & \textbf{5.41} & 0.16 & 5.09 & \textbf{0.00} & 0.20 & \textbf{95.46} \\
    & TemporalMaxer             & 86.52 & 83.37 & 83.66 & 0.41 & 6.02 & 0.19 & 4.41 & \textbf{0.00} & 0.24 & 94.39 \\
    & TriDet                    & \textbf{88.95} & \textbf{86.15} & \textbf{86.45} & 0.38 & \textbf{5.41} & \textbf{0.12} & \textbf{3.95} & 0.01 & 0.29 & 94.75 \\
    \hline
    \multirow{7}{*}{\rotatebox[origin=c]{90}{Wetlab}}
    & DeepConvLSTM              & 38.65 & 47.34 & 37.87 & \textbf{0.51} & \textbf{7.13} & 0.69 & 48.53 & 0.21 & 0.64 & 11.88 \\
    & Shallow D.                & 39.01 & 35.42 & 34.42 & \textbf{0.51} & 9.52 & 1.57 & \textbf{34.51} & \textbf{0.06} & \textbf{0.43} & 15.40 \\
    & A-and-D                   & 37.75 & \textbf{55.71} & 37.49 & 0.56 & 9.69 & \textbf{0.63} & 57.60 & 0.16 & 0.57 & 12.27 \\
    & TinyHAR                   & 34.31 & 50.84 & 31.48 & 0.61 & 8.41 & 1.00 & 61.26 & 0.11 & 0.59 & 10.05 \\ \cline{2-12}
    & ActionFormer              & 40.71 & 49.25 & 40.71 & 0.56 & 9.41 & 0.79 & 52.44 & 0.07 & 0.79 & 33.53 \\
    & TemporalMaxer             & \textbf{50.43} & 36.65 & 37.09 & 0.59 & 9.37 & 0.83 & 53.59 & 0.10 & 0.61 & \textbf{35.72} \\
    & TriDet                    & 44.13 & 49.15 & \textbf{42.85} & 0.58 & 8.85 & 0.75 & 48.63 & 0.10 & 0.84 & 34.05 \\
    \hline
    \multirow{7}{*}{\rotatebox[origin=c]{90}{WEAR}}
    & DeepConvLSTM              & 80.68 & 76.25 & 75.78 & 0.28 & \textbf{2.35} & 0.52 & 6.68 & 0.11 & \textbf{0.32} & 61.03 \\
    & Shallow D.                & 80.78 & 78.91 & 77.71 & 0.27 & 3.23 & 0.50 & \textbf{5.21} & 0.04 & 0.43 & 67.89 \\
    & A-and-D                   & \textbf{82.34} & 83.29 & \textbf{80.61} & \textbf{0.20} & 4.03 & 0.33 & 7.18 & 0.09 & 0.52 & 64.78 \\
    & TinyHAR                   & 81.87 & \textbf{84.23} & 80.56 & 0.21 & 4.77 & \textbf{0.27} & 9.54 & 0.10 & 0.55 & 63.33 \\ \cline{2-12}
    & ActionFormer              & 71.88 & 76.70 & 72.43 & 0.22 & 6.48 & 0.65 & 7.12 & \textbf{0.01} & 2.06 & 73.80 \\
    & TemporalMaxer             & 69.54 & 72.80 & 69.52 & 0.23 & 6.87 & 0.83 & 5.50 & \textbf{0.01} & 1.50 & 69.18 \\
    & TriDet                    & 73.57 & 77.54 & 73.18 & 0.28 & 4.79 & 0.64 & 6.21 & 0.03 & 1.64 & \textbf{77.12} \\
    \hline
    \multirow{7}{*}{\rotatebox[origin=c]{90}{Hang-Time}}
    & DeepConvLSTM              & 44.13 & 33.95 & 35.25 & \textbf{0.28} & 10.60 & 0.88 & \textbf{20.16} & 0.27 & 1.46 &  5.44 \\
    & Shallow D.                & 37.97 & 38.19 & 36.85 & 0.35 & 14.00 & 1.07 & 36.38 & \textbf{0.21} & 2.33 &  5.00 \\
    & A-and-D                   & 40.54 & 43.32 & 40.39 & 0.35 & 14.14 & 0.72 & 34.81 & 0.30 & 1.44 &  6.73 \\
    & TinyHAR                   & 37.09 & 41.13 & 36.89 & 0.41 & 11.59 & 0.75 & 42.90 & 0.43 & 1.26 &  4.73 \\ \cline{2-12}
    & ActionFormer              & 49.19 & \textbf{57.57} & \textbf{51.23} & 0.62 & 11.63 & 0.51 & 47.65 & 0.48 & 0.64 & \textbf{29.26} \\
    & TemporalMaxer             & 45.01 & 54.56 & 47.17 & 0.71 & 10.97 & \textbf{0.45} & 52.71 & 1.13 & 0.65 & 27.86 \\
    & TriDet                    & \textbf{49.59} & 55.14 & 50.67 & 0.73 & \textbf{9.85} & 0.52 & 48.55 & 0.69 & \textbf{0.62} & 29.24 \\
    \hline
    \multirow{7}{*}{\rotatebox[origin=c]{90}{RWHAR}}
    & DeepConvLSTM              & 79.05 & 81.93 & 77.56 & 0.65 & 2.05 & 1.09 & 13.63 & 1.07 & \textbf{0.00} &  0.11 \\
    & Shallow D.                & \textbf{88.59} & \textbf{89.01} & \textbf{86.85} & \textbf{0.31} & \textbf{1.14} & \textbf{0.38} & \textbf{6.93} & 0.98 & \textbf{0.00} & 0.00 \\
    & A-and-D                   & 79.46 & 82.68 & 78.09 & 0.66 & 2.31 & 1.17 & 11.28 & 0.84 & \textbf{0.00} &  0.04 \\
    & TinyHAR                   & 83.59 & 86.25 & 82.62 & 0.53 & \textbf{1.14} & 0.76 & 11.65 & 0.89 & \textbf{0.00} & 0.02 \\ \cline{2-12}
    & ActionFormer              & 63.76 & 67.64 & 61.24 & 2.48 & 11.12 & 2.06 & 11.23 & \textbf{0.09} & \textbf{0.00} & 65.40 \\
    & TemporalMaxer             & 63.20 & 67.60 & 60.59 & 2.59 & 11.95 & 1.81 & 13.96 & 0.36 & 0.05 & 50.60 \\
    & TriDet                    & 69.27 & 73.04 & 67.86 & 1.48 & 6.88 & 2.03 & 10.24 & 0.23 & \textbf{0.00} & \textbf{69.98} \\
    \hline
  \end{tabular}
\end{table*}

\begin{figure*}
\centering
\includegraphics[width=.9\textwidth]{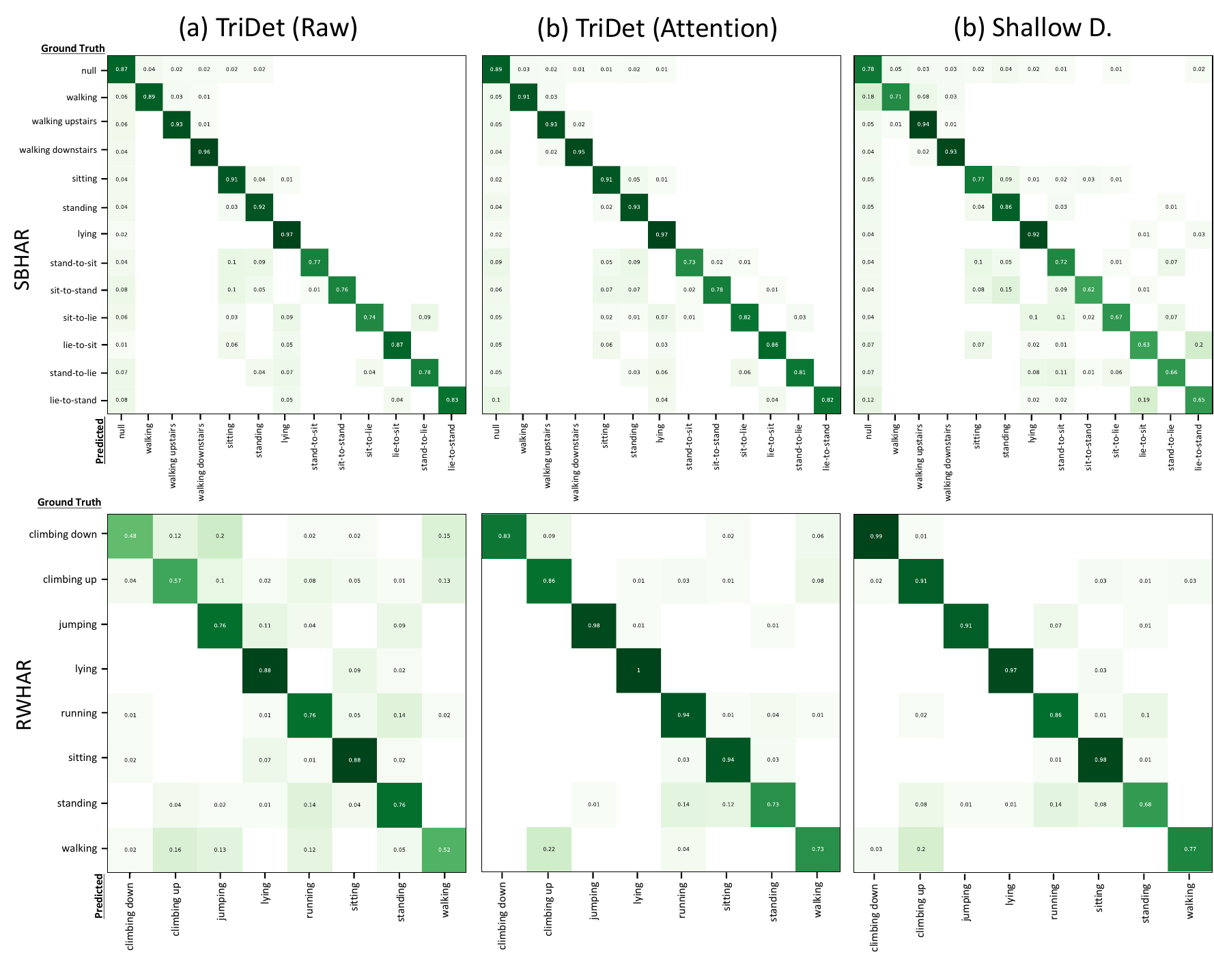}
    \caption{\textit{Offline Activity Recognition:} Confusion matrices of the (a) the best TAL architecture (TriDet) \cite{shiTriDetTemporalAction2023} and (b) inertial model (shallow DeepConvLSTM) being applied on the SBHAR \cite{reyes-ortizTransitionAwareHumanActivity2016} (top) and RWHAR dataset \cite{sztylerOnbodyLocalizationWearable2016} (bottom) with a one second sliding window and 50\% overlap. Note that confusions which are 0 are omitted.}
    \label{fig:sbharrwhar}
    \Description{\textit{Offline Activity Recognition:} Confusion matrices of the (a) the best TAL architecture (TriDet) and (b) inertial model (shallow DeepConvLSTM) being applied on the SBHAR \cite{reyes-ortizTransitionAwareHumanActivity2016} (top) and RWHAR dataset (bottom) with a one second sliding window and 50\% overlap. Note that confusions which are 0 are omitted.}
\end{figure*}

With the improvements on the SBHAR being the largest across all datasets, one can see on a per class level (see Figure \ref{fig:sbharrwhar}) that this increase can be attributed to improved performance on transitional, non-periodic classes like \textit{sit-to-stand}. These activities are mostly recognisable by their context and surrounding activities and are thus particularly challenging to predict for models that do not rely on temporal dependencies spanning multiple seconds. 
By applying the TAL architectures in an offline manner, the architectures are able to leverage both local and global context across the whole timeline and are thus even able to recognize these short-lasting activities.
Since the DeepConvLSTM, Attend-and-Discriminate and TinyHAR models are being trained on shuffled training data and have recurrent parts applied on within-window sequences, the three architectures cannot learn to predict these activities based on surrounding context. Among inertial models, only the shallow DeepConvLSTM is able to more reliably predict these context-based classes as it is trained on unshuffeled data and applies a dimensional flip when training the LSTM. Across all datasets part of this benchmark analysis, the RWHAR dataset yields the least performant results for the TAL architectures. We accredit this primarily due to the absence of a NULL-class in the dataset, which introduces a uncommon prediction scenario for the models. Furthermore, the RWHAR consists of the least amount of segments per subject, limiting the amount of training segments which can be used to optimize the TAL models. Results on the RWHAR dataset are nevertheless surprising as the TAL models show confusion among classes which they do not struggle to predict in other datasets (e.g., \textit{lying} in the SBHAR dataset) as well as classes that are not similar to each other (e.g., \textit{lying} and \textit{jumping}). The obtained results raise the question of whether TAL models are primarily suited for being applied on untrimmed sequences, which (1) include breaks and/ or (2) provide a larger amount of segments than the RWHAR dataset. 

Nevertheless, apart from the RWHAR dataset, the TAL models deliver the most consistent results across all classes. Even though, in most cases, the individual per-class accuracies are not the highest when compared to results obtained using the inertial architectures, the inertial architectures are frequently not able to predict all classes reliably, with at least one class showing low prediction accuracy. This is especially evident when examining the results obtained on the Hang-Time and Opportunity datasets, where TAL models are capable of correctly predicting challenging non-periodic activities such as \textit{rebounds}, \textit{passes}, \textit{opening door} and \textit{closing door}. Furthermore, as also seen in the DETAD analysis, TAL models are overall more capable of differentiating activities from the NULL-class, showing the highest NULL-class accuracy across all datasets. To summarize, the TAL architectures are more reliable in recognizing any kind of actions within an untrimmed sequence, and are less prone to predict fragmented prediction streams or non-existent breaks.

\begin{figure*}
\centering
\includegraphics[width=.9\textwidth]{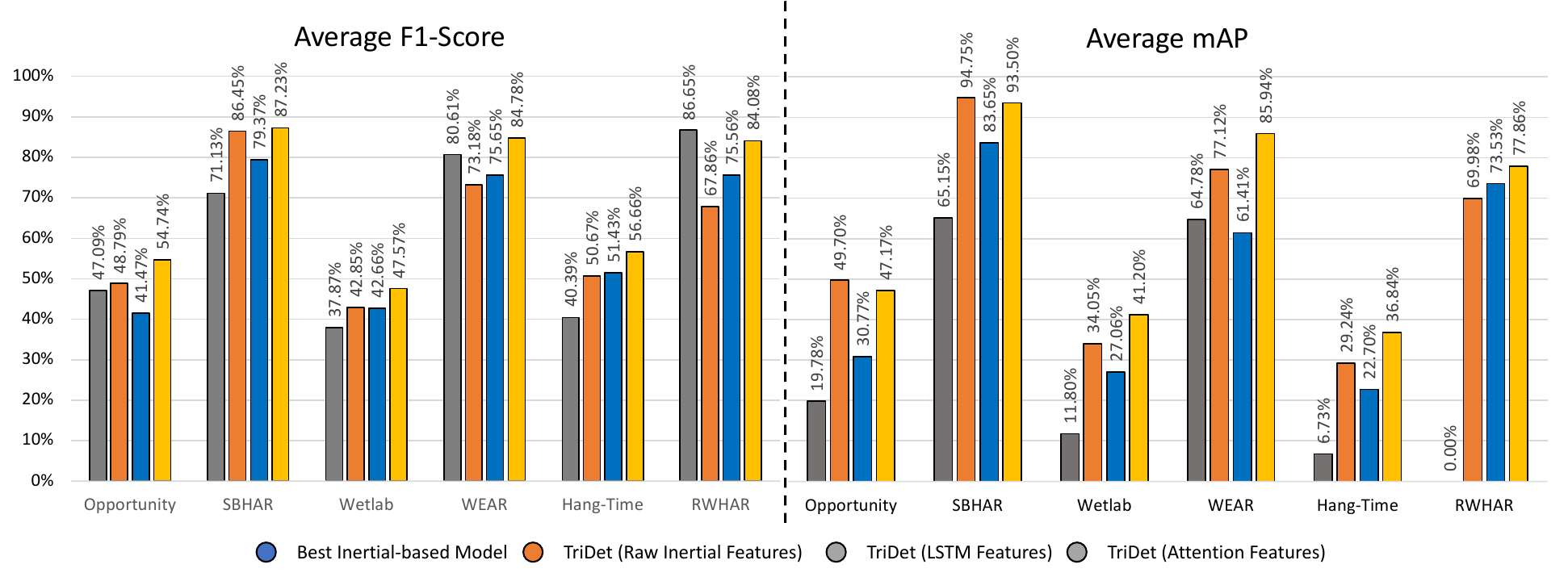}
    \caption{\textit{Offline Activity Recognition}: Average LOSO results of the two-stage TAL training of the TriDet model. We compare the LSTM and attention feature-based two-stage TAL training with the best inertial-based architecture and the single-stage TAL training on raw inertial data of all six benchmark datasets. One can see a clear improvement when using attention-based features across all datasets.}
    \Description{\textit{Offline Activity Recognition}: Average LOSO results of the two-stage TAL training of the TriDet model. We compare the LSTM and attention feature-based two-stage TAL training with the best inertial-based architecture and the single-stage TAL training on raw inertial data of all six benchmark datasets. Results are presented as bar plots. One can see a clear improvement when using attention-based features across all datasets.}
    \label{fig:two_stage_results}
\end{figure*}

\paragraph{Single- vs. Two-Stage TAL Training}
As described in Section \ref{sec:talforhar}, the TAL models described in this paper were intended to be applied using feature embeddings describing each sliding window, rather than raw data. Nevertheless, our initial results show that TAL models are indeed capable of being applied to raw inertial data. Figure~\ref{fig:two_stage_results} presents results of our implemented two-stage TAL training, which extends the training process as described in the previous chapters with a prepended feature extraction using inertial models. In total, we assess two feature embeddings: LSTM-based features extracted from a DeepConvLSTM \cite{ordonezDeepConvolutionalLSTM2016} and attention-based features from a TinyHAR architecture \cite{zhouTinyHARLightweightDeep2022}. Though the two-stage training using LSTM-based features only yields better F1-scores for the Wetlab, WEAR, and RWHAR datasets, the training using attention-based features improves results across all datasets across all TAL models. Given the significant improvements on the RWHAR and WEAR datasets, we assume that the feature extraction via the prepended inertial network helps increase the discriminability of the window-level features, yet at the cost of making it harder for the TAL models to learn cross-window temporal relations, as evident by the decrease in mAP scores across (almost) all datasets.

\subsection{Online Activity Recognition}

\begin{table*}
  \centering
  \scriptsize
  \caption{Average LOSO results obtained using the evaluated TAL architectures \cite{zhangActionFormerLocalizingMoments2022, tangTemporalMaxerMaximizeTemporal2023, shiTriDetTemporalAction2023} employing different degrees of chunking during validation. During validation each subject is split into equal-sized chunks with each chunk being individually predicted by the trained TAL models. The table provides per-sample F1-Score (F1), average mAP applied at different tIoU thresholds (0.3:0.1:07) calculated on each chunk individually (c-mAP) and on the reconstructed validation stream (mAP). Results are averaged across three runs employing different random seeds. Results which are \ul{underlined} are not score thresholded.}
  \label{tab:chunkedresults}
{
  \begin{tabular}{@{}ll|ccc|ccc|ccc|ccc|cc@{}}
    & & \multicolumn{3}{c}{1 sec.} & \multicolumn{3}{c}{5 sec.} & \multicolumn{3}{c}{30 sec.} & \multicolumn{3}{c}{60 sec.} & \multicolumn{2}{c}{Unchunked} \\ 
    & Model & F1 & c-mAP & mAP & F1 & c-mAP & mAP & F1 & c-mAP & mAP & F1 & c-mAP & mAP & F1 & mAP \\ 
    \hline
    \multirow{3}{*}{\rotatebox[origin=c]{90}{Opp.}}
    & AF             & \ul{2.63} & \ul{2.45} & \ul{0.36} & 37.73 & 18.10 & 25.51 & 49.96 & 41.74 & 45.84 & 50.50 & 45.45 & 47.84 & 51.93 & 51.24 \\
    & TM             & \ul{2.22} & \ul{2.09} & \ul{0.31} & 38.00 & 19.08 & 25.99 & 45.32 & 39.24 & 42.83 & 45.27 & 42.15 & 44.22 & 44.74 & 46.31 \\
    & TD             & \ul{1.48} & \ul{2.26} & \ul{0.37} & \ul{9.96} & \ul{10.51} & \ul{13.6} & \ul{12.15} & \ul{12.73} & \ul{13.37} & \ul{1.99} & \ul{4.15} & \ul{4.27} & 48.79 & 49.70 \\  [0.5ex] 
    \hline
    \multirow{3}{*}{\rotatebox[origin=c]{90}{SBHAR}}
    & AF             & \ul{3.44} & \ul{7.56} & \ul{0.12} & \ul{34.49} & \ul{26.20} & \ul{15.09} & 78.88 & 74.28 & 79.52 & 81.57 & 82.45 & 86.39 & 84.43 & 95.46 \\ 
    & TM             & \ul{3.30} & \ul{5.39} & \ul{0.05} & \ul{33.49} & \ul{47.11} & \ul{23.87} & 78.08 & 73.95 & 77.64 & 81.04 & 82.41 & 85.19 & 83.66 & 94.39 \\
    & TD             & \ul{2.13} & \ul{10.18} & \ul{0.13} & \ul{17.69} & \ul{29.31} & \ul{8.05} & \ul{31.70} & \ul{52.01} & \ul{49.38} & \ul{44.76} & \ul{72.94} & \ul{72.64} & 86.45 & 94.75 \\  [0.5ex]
    \hline
    \multirow{3}{*}{\rotatebox[origin=c]{90}{Wetlab}}
    & AF             & \ul{5.13} & \ul{2.37} & \ul{0.02} & \ul{14.55} & \ul{3.12} & \ul{0.77} & 35.35 & 25.65 & 20.15 & 35.98 & 30.46 & 28.06 & 40.71 & 33.53 \\ 
    & TM             & \ul{4.82} & \ul{2.23} & \ul{0.03} & \ul{11.33} & \ul{15.81} & \ul{1.51} & 34.37 & 25.87 & 20.16 & 35.35 & 30.24 & 26.90 & 37.09 & 35.72 \\
    & TD             & \ul{1.77} & \ul{4.08} & \ul{0.03} & \ul{2.20} & \ul{4.40} & \ul{0.63} & \ul{4.03} & \ul{6.10} & \ul{7.38} & \ul{7.78} & \ul{10.98} & \ul{9.11} & 42.85 & 34.05 \\ [0.5ex] 
    \hline
    \multirow{3}{*}{\rotatebox[origin=c]{90}{WEAR}}
    & AF             & \ul{5.70} & \ul{3.47} & \ul{0.00} & \ul{15.54} & \ul{10.44} & \ul{0.06} & 53.57 & 40.02 & 18.72 & 69.21 & 56.89 & 52.11 & 72.43 & 73.80 \\
    & TM             & \ul{5.30} & \ul{2.65} & \ul{0.00} & \ul{38.37} & \ul{53.39} & \ul{0.19} & 58.29 & 45.09 & 18.94 & 67.91 & 56.60 & 47.89 & 69.52 & 69.18 \\
    & TD             & \ul{1.25} & \ul{5.41} & \ul{0.00} & \ul{3.72} & \ul{15.12} & \ul{0.09} & \ul{34.67} & \ul{49.73} & \ul{21.57} & \ul{43.96} & \ul{49.33} & \ul{43.79} & 73.18 & 77.12 \\ [0.5ex] 
    \hline
    \multirow{3}{*}{\rotatebox[origin=c]{90}{Hang-T.}}
    & AF             & \ul{6.41} & \ul{3.19} & \ul{0.70} & 42.41 & 17.18 & 14.67 & 49.32 & 26.47 & 26.82 & 50.69 & 27.64 & 27.65 & 51.23 & 29.26 \\
    & TM             & \ul{6.39} & \ul{2.63} & \ul{0.50} & 41.13 & 17.54 & 15.17 & 48.15 & 25.72 & 25.38 & 48.67 & 26.63 & 26.33 & 47.17 & 27.86 \\
    & TD             & 19.52 & 0.00 & 3.56 & 19.33 & 3.67 & 3.48 & 26.02 & 12.42 & 13.01 & 35.72 & 18.80 & 19.15 & 50.67 & 29.24 \\ [0.5ex]
    \hline
    \multirow{3}{*}{\rotatebox[origin=c]{90}{RWHAR}}
    & AF             & \ul{8.47} & \ul{28.67} & \ul{0.00} & \ul{17.00} & \ul{51.96} & \ul{0.00} & \ul{50.15} & \ul{76.19} & \ul{1.37} & \ul{62.40} & \ul{77.58} & \ul{7.91} & \ul{61.24} & \ul{65.40} \\ 
    & TM             & \ul{8.06} & \ul{17.13} & \ul{0.00} & \ul{31.35} & \ul{48.89} & \ul{0.00} & \ul{52.71} & \ul{60.47} & \ul{1.31} & \ul{58.47} & \ul{61.86} & \ul{7.61} & \ul{60.59} & \ul{50.60} \\
    & TD             & \ul{5.88} & \ul{25.68} & \ul{0.00} & \ul{9.07} & \ul{40.11} & \ul{0.00} & \ul{39.78} & \ul{47.78} & \ul{0.94} & \ul{48.65} & \ul{57.51} & \ul{7.39} & \ul{67.86} & \ul{69.98} \\ [0.5ex]
    \hline
  \end{tabular}
}
\end{table*}

A major difference in how TAL models are intended to be used compared to inertial models is that they can only provide an offline prediction of previously recorded timeseries data. In previous experiments, we demonstrated that TAL models are capable of analyzing a participant's data stream as a whole and can even deal with arbitrary lengths of timeseries data and activity durations. However, to investigate how capable TAL models are at analyzing only small snippets of incoming data streams, we modified the previously employed Leave-One-Subject-Out validation loop so that TAL models are tasked to predict each validation subjects' data in a chunked manner. Specifically, given the data of a validation participant, we divided the data into equal-sized portions (e.g. 5 seconds worth of data) and had the TAL models, trained using unchunked data, predict each chunk individually. Table \ref{tab:chunkedresults} compares the initial unchunked results of the tested TAL models (ActionFormer, TemporalMaxer and TriDet) on the six inertial benchmark datasets with results obtained when fragmenting each validation split into 1, 5, 30, or 60-second chunks. Across all datasets one can witness that with smaller-sized chunks classification and mAP results worsen across all algorithms and datasets. Overall, we can see that the smaller the chunk, the lower the overall confidence of the TAL models regarding each predicted segment. The lower confidence further leds to almost all segment predictions being removed even with a low score threshold such as 0.05, resulting in us not applying score thresholding in those cases, i.e., using all segments predicted by the respective TAL model. We assume that this drop in prediction confidence is likely caused by artificially splitting long segments of activities, contradicting what the models have seen during training, as datasets such as the Opportunity and Hang-Time dataset, which contain mostly short-lasting segments, are less affected by the smaller chunks as e.g. the WEAR dataset which has almost exclusively segments lasting longer than 30 seconds. While the TriDet model is not capable of dealing with the chunked validation, most likely due to it using a different style of regression head, the ActionFormer and the TemporalMaxer are capable of maintaining predictive performance even for small-sized chunks of 5 seconds across almost all datasets. As expected, mAP scores are more impacted by the chunked prediction as chances are increased that long lasting segments are split due to the predicted segments not spanning across chunks. Nevertheless, we expect this effect to dampen if one would use an additional majority filtering as used in our inertial-based experiments, as this would eliminate potential intermediate activity switches.

Table~\ref{tab:inference} provides a comparison in terms of learnable parameters, size on disk, average batch training and inference times of the inertial-based models compared with the TAL models. One can see that the TAL models are significantly larger in size and number of learnable parameters. Though all architectures part of our analysis are capable of predicting a window of one second as well as the complete prediction stream of a participant within less than half a second, inertial-based models are faster than TAL models in both inference scenarios being on average faster by around a magnitude of 10. Taking into consideration that all TAL models fail to predict 1 second chunks and are significantly larger than inertial-based models, it becomes aparent, that TAL models are not suited for inference on edge-devices. It becomes apparent that with the models being trained to predict activity occurrences as a whole, the models require enough surrounding context in order to spot activities and correctly identify what they have seen during training. Nevertheless, as soon as the TAL models are given enough surrounding context (e.g., a prior context of 60 seconds), the models are capable to spot activities within said activity stream in a reasonable run time.

\begin{table*}
  \centering
  \scriptsize
  \caption{Comparison in terms of learnable parameters (in million (M)), size on disk (in MB), average batch training and inference times (in milliseconds) of the seven benchmark algorithms. Training and inference speeds are the average across the first 5 epochs of the first LOSO validation split of the algorithms being applied on the WEAR dataset. Benchmarking was performed on a single NVIDIA GeForce RTX 4090 with an AMD Ryzen 7800 X3D. We assess both inference speeds of the architectures being tasked to predict one 1-second window as well as one complete subject. $^{**}$ 1 batch equals 100 1-second windows $^{**}$ 1 batch equals 1 participant.}
  \label{tab:inference}
{
  \begin{tabular}{@{}llllllll@{}}
    & Model & Total Params & Size & \multicolumn{3}{c}{Avg. Time per Batch} \\ \cmidrule{5-7}
    &  &  &  & Train & Test (1 sec.) &  Test (1 participant) \\ 
    \hline
    \multirow{7}{*}{\rotatebox[origin=c]{90}{Baseline}}
    & DeepConvLSTM              & 0.71 M & 2.69MB & 4ms$^*$   & 2ms  & 27ms \\
    & Shallow D.                & 0.57 M & 2.19MB & 3ms$^*$   & 2ms  & 29ms \\
    & A-and-D                   & 0.57 M & 2.19MB & 31ms$^*$  & 10ms & 41ms \\
    & TinyHAR                   & 0.04 M & 0.14MB & 7ms$^*$   & 3ms  & 28ms \\
    & ActionFormer              & 27.02 M & 103.37MB & 131ms$^{**}$ & 39ms & 346ms \\
    & TemporalMaxer             & 4.89 M & 18.94MB & 69ms$^{**}$  & 25ms & 380ms \\
    & TriDet                    & 13.75 M & 52.73MB & 127ms$^{**}$ & 37ms & 347ms \\ 
    \hline
  \end{tabular}
}
\end{table*}

\begin{figure*}
\centering
\includegraphics[width=.7\textwidth]{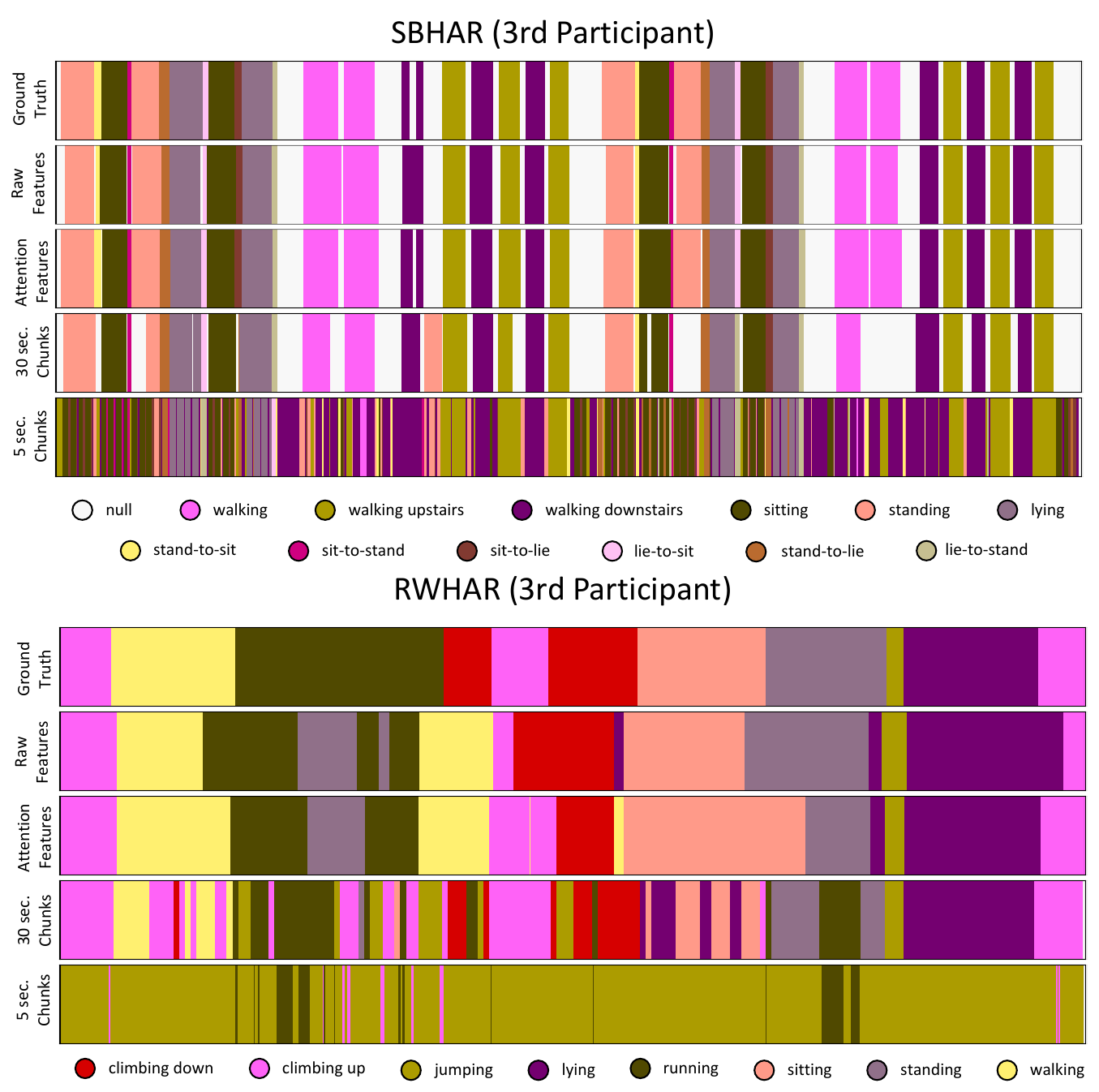}
    \caption{Color-coded visualization of the effect of the \textit{Offline} and \textit{Online Activity Recognition} experiments using the ActionFormer model \cite{zhangActionFormerLocalizingMoments2022}. Results are compared with the ground truth of a sample participant as well as offline single-stage training for both the RWHAR \cite{sztylerOnbodyLocalizationWearable2016} and SBHAR \cite{reyes-ortizTransitionAwareHumanActivity2016} dataset. One can see that training the TAL models using attention based features clearly improves results with an overall better recognition of all activities. Contrarily, once predicting a chunked output, performance can be somewhat maintained for larger chunks, yet drops near zero once applied on only 5-second long windows.}
    \label{fig:bitmapviz}
    \Description{Visualization of the prediction streams of the ActionFormer for single-stage, two-stage and chunked prediction scenarios. Results are visualized for both the SBHAR and RWHAR dataset and compared against ground truth of a sample participant.}
\end{figure*}

\section{Discussion \& Conclusion}
This article demonstrated the applicability of vision-based, single-stage Temporal Action Localization for inertial-based Human Activity Recognition. We showed that three state-of-the-art TAL models \cite{zhangActionFormerLocalizingMoments2022, tangTemporalMaxerMaximizeTemporal2023, shiTriDetTemporalAction2023} can be applied in a plain fashion to raw inertial data and achieve competitive results on six popular inertial HAR datasets \cite{roggenCollectingComplexActivity2010, reyes-ortizTransitionAwareHumanActivity2016, schollWearablesWetLab2015, sztylerOnbodyLocalizationWearable2016, hoelzemannHangtimeHARBenchmark2023, bockWEAROutdoorSports2023}, outperforming in most cases popular models from the inertial community in an offline prediction scenario by a significant margin. Using a combination of an inertial network as a feature extractor and a TAL model showed to significantly enhance classification results for all TAL models -- especially on datasets where classic inertial models had previously outperformed TAL models. Our two-stage experiments further suggest potential improvements that could be achieved by investigating improved methods for feature extraction using TAL models and inertial data, e.g. via a fully-differentiable combined version of both type of architectures (see Figure~\ref{fig:bitmapviz}).

A previously unexplored metric in inertial-based HAR, mean Average Precision (mAP), reveals that TAL models predict less fragmented timelines compared to inertial models and overall achieve larger degrees of overlap with ground truth segments. Furthermore, TAL models show to predict even non-periodic and complex activities more reliably than inertial architectures, providing consistent results across all types of classes across all datasets. Being one of the key challenges in HAR \cite{bullingTutorialHumanActivity2014}, the TAL architectures are further shown to be less affected by the unbalanced nature of HAR datasets due to a large NULL-class. Across the five benchmark datasets which offered a NULL-class, the TAL architectures showed to deliver the highest NULL-class accuracy. Additional experiments, which involved applying the TAL models on artificially created chunked sequences of data, showed that TAL models, though intended for the offline analysis of prerecorded activity timelines, have the capability of being applied in a near-online fashion (see Figure~\ref{fig:bitmapviz}). Although the TAL models highlighted in this paper overall size and complexity would not allow them to be run on edge devices (yet), their reasonable performance and inference time on large enough chunks suggest that they could function e.g. as a server-side prediction model as seen in \cite{berchtoldActiServActivityRecognition2010}, which analyzes chunks of data in regular intervals.

The research community for inertial-based activity recognition has contributed methods to better model temporal relationships in the past years, yet most such architectures are limited to learning context within a fixed-sized sliding window. To this date, the length of the sliding window remains a crucial parameter in inertial-based HAR, which may result in a significant performance drops in recognition performance when set incorrectly. Specifically, if dealing with both long and short lasting activities, a small sliding window might end up being too small to fully capture an activity, while a too large window size might cause sliding windows containing mixed activity types. With studies such as \citet{pellattCausalBatchSolvingComplexity2020} and \citet{guanEnsemblesDeepLSTM2017} contributing strategies to improve training, the TAL community offers an interesting new perspective to inertial-based HAR -- alleviating the need for fixed size windows and making algorithms capable of dealing with activities of arbitrary length. Although models from these two independent communities share similarities, the TAL community offers many unexplored design choices and training concepts for a multitude of application scenarios, which we argue should be considered for investigation by the inertial-based HAR community.

\begin{acks}
We gratefully acknowledge the DFG Project WASEDO (grant number 506589320) and the University of Siegen's OMNI cluster.
\end{acks}

\bibliographystyle{ACM-Reference-Format}
\bibliography{main}

\end{document}